\documentclass[10pt,twocolumn,letterpaper]{article}

\usepackage{cvpr} % uncomment this for the final submission
\usepackage{times}
\usepackage{epsfig}
\usepackage{graphicx}
\usepackage{amsmath}
\usepackage{amssymb}

\usepackage[table]{xcolor} % Allows cell coloring
\usepackage{colortbl}      % Additional table coloring support
\usepackage{booktabs}      % For clean table formatting
\usepackage{makecell}
\usepackage{multirow}
\usepackage{placeins}

\usepackage{float}
\usepackage{dblfloatfix}

\usepackage[pagebackref=true,breaklinks=true,letterpaper=true,colorlinks,bookmarks=false]{hyperref}

\begin{document}

%%%%%%%%% TITLE
\title{Saliency-Guided Training for Fingerprint Presentation Attack Detection}

% \author{Samuel Webster \\
% % University of Notre Dame\\
% {\tt\small swebster@nd.edu}
% \and
% Adam Czajka\\
% University of Notre Dame\\
% {\tt\small aczajka@nd.edu}
% }

\author{Samuel Webster\hspace{2cm}Adam Czajka\\
Department of Computer Science and Engineering\\
University of Notre Dame, IN, USA\\
{\tt\small \{swebster,aczajka\}@nd.edu}
}

\newcommand{\teaser}{
{
    \vskip3mm
    \begin{center}
    \vskip0mm\includegraphics[width=\linewidth]{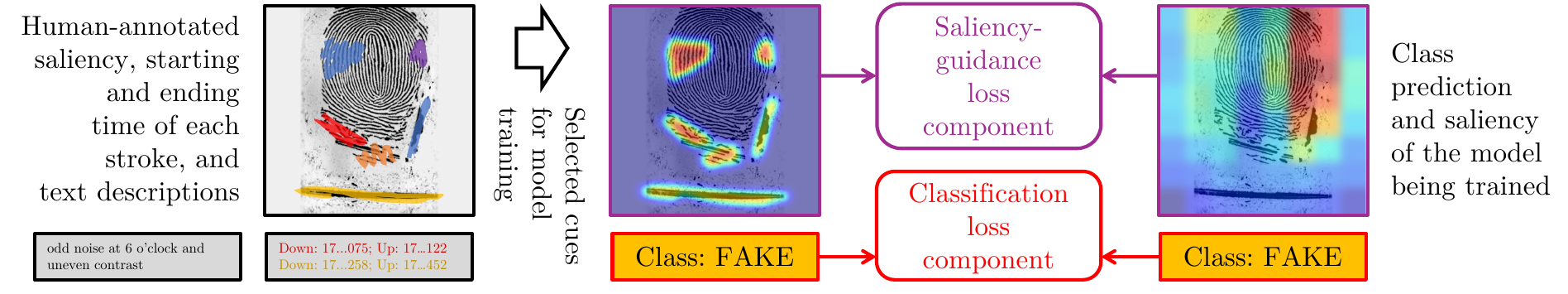}
    
    % Adam's version
    \captionof{figure}{We collected visual salient annotations and text descriptions, supporting decisions of humans detecting fake fingerprints, and used these salient features in saliency-guided model training paradigm to offer an effective fingerprint presentation attack detection method.}
    \label{fig:teaser}
    \end{center}
}
\vspace*{-10pt}
}

\maketitle
\thispagestyle{empty}

%%%%%%%%% ABSTRACT
\begin{abstract}
    \vspace*{-5pt}
    Saliency-guided training, which directs model learning to important regions of images, has demonstrated generalization improvements across various biometric presentation attack detection (PAD) tasks. This paper presents its first application to fingerprint PAD. We conducted a 50-participant study to create a dataset of 800 human-annotated fingerprint perceptually-important maps, explored alongside algorithmically-generated ``pseudosaliency,'' including minutiae-based, image quality-based, and autoencoder-based saliency maps. Evaluating on the 2021 Fingerprint Liveness Detection Competition testing set, we explore various configurations within five distinct training scenarios to assess the impact of saliency-guided training on accuracy and generalization. Our findings demonstrate the effectiveness of saliency-guided training for fingerprint PAD in both limited and large data contexts, and we present a configuration capable of earning the first place on the LivDet-2021 benchmark. Our results highlight saliency-guided training's promise for increased model generalization capabilities, its effectiveness when data is limited, and its potential to scale to larger datasets in fingerprint PAD.
    All collected saliency data and trained models are released with the paper to support reproducible research.
\end{abstract}

%%%%%%%%% BODY TEXT

% Introduction
\section{Introduction}
\label{sec:Introduction}

Fingerprint presentation attack detection (PAD) is an important challenge in biometrics, aimed at distinguishing between genuine fingerprint samples and spoofed (fake) ones. Presentation attacks refer to attempts to deceive fingerprint recognition systems using artificial means, such as synthetic fingerprint impressions made from materials like silicone or gel. Given that fingerprint biometrics are widely employed in systems requiring high levels of security, enhancing their ability to detect spoofing attempts is essential. Deep learning-based fingerprint PAD systems, while effective in detecting known attack types, struggle with generalizing to new or unseen spoofing techniques \cite{karampidis2021comprehensive}. These systems are further limited by the scarcity of rarer spoofs, reducing the diversity of known attack types and impacting the ability to evaluate novel ones. This lack of generalization remains a critical issue as fingerprint attack methods continue to evolve rapidly \cite{sousedik2014presentation}. 

Saliency-guided training offers a promising solution to the challenges of fingerprint PAD by directing model decision-making to salient regions, focusing learning on discriminative features and on ignoring features accidentally correlated with class labels. Saliency guidance has been applied through both image transformation-based and loss-based approaches, including blurring non-salient regions \cite{Boyd2021} and aligning Class Activation Mapping (CAM) \cite{zhou2016learning} with salient regions \cite{boyd2023cyborg}, respectively. This technique has been shown to improve model training and convergence, particularly in scenarios with limited data \cite{Linsley2019}. Saliency-guided training can also improve model trustworthiness, fairness, and explainability  \cite{crum2024taking}.

This paper aims to investigate whether saliency-guided training can improve fingerprint PAD performance by directing model learning to salient fingerprint regions. To explore this intersection, we introduce a dataset of 800 doubly human-annotated saliency maps and text descriptions collected in a 50-participant study, using bonafide and spoof fingerprint samples sourced from previous Fingerprint Liveness Detection Competitions (LivDet) \cite{livdet15, livdet17, livdet19, livdet21}. Additionally, we introduce algorithmically-sourced \textit{pseudosaliency} maps, including autoencoder-generated saliency, fingerprint matcher-sourced minutiae-based saliency, and fingerprint image quality-based saliency.

We use all saliency types introduced above in a saliency-guided training paradigm to train fingerprint PAD models and answer the following research questions:

\begin{enumerate}
    \setlength\itemindent{28.5pt}
    \item[\bf -- RQ1:] Does saliency-guided training with human-annotative saliency improve generalization and classification accuracy for fingerprint PAD? %We hypothesize that saliency guidance will improve fingerprint PAD performance due to the shared problems it solves in other biometric modalities.
    \item[\bf -- RQ2:] Are algorithmically-sourced pseudosaliency types effective for fingerprint PAD? % We hypothesize that non-directly-human-sourced pseudosaliency will match or exceed the performance of human-annotative saliency based on improvements seen in other biometric modalities.
    \item[\bf -- RQ3:] How does saliency-guided training affect fingerprint PAD generalization compared to other biometric attack detection tasks? % We hypothesize that the impact of saliency-guided training on fingerprint PAD performance will be comparable to those observed in other biometric modalities (iris and face PAD). However, we recognize that fingerprint PAD systems generally exhibit higher baseline performance than iris PAD or synthetic face detection tasks, which could influence the relative gain we may achieve by supplementing the training with saliency-dependent components.
\end{enumerate}

Alongside contributions at the intersection of saliency-guided training and fingerprint PAD, this paper releases all explored saliency data at \texttt{https://github.com/CVRL/Human-Machine-\\Pairing-Fingerprint-PAD}, including the 800 doubly human-annotated saliency maps, the various sets of algorithmically-sourced pseudosaliency and the trained fingerprint PAD models.

% \footnote{The link to a GitHub repository, with all components and instructions on how to request a copy of the dataset, has been removed for anonymity and will be added in the event of this paper being accepted for presentation at IJCB 2025.} 

% Literature Review
\section{Related Work}

Presentation attack detection is a lasting challenge within many biometric modalities, including iris \cite{czajka2018presentation}, face \cite{ramachandra2017presentation, bhattacharjee2019recent}, voice \cite{sahidullah2023introduction}, or fingerprint \cite{sousedik2014presentation, karampidis2021comprehensive} recognition systems. Saliency-guided training has been reported to increase generalization in a variety of biometric tasks: iris PAD \cite{Boyd2021, crum2023teaching, Crum2024}, synthetic face detection \cite{boyd2023cyborg, Crum2023, crum2023teaching, Crum2024}, and chest X-ray analysis \cite{Vansonsbeek2022, Wang2023}. This section, due to the paper's goal related to exploration of saliency-based guidance of fingerprint PAD models, will discuss shortly the relevant literature on neural networks-based fingerprint PAD, and applications of saliency in training deep learning models solving biometric PAD tasks. 
%\begin{enumerate}
%    \item \textbf{Neural Networks for Fingerprint PAD}: An overview of deep learning approaches to fingerprint PAD, including patch-based analysis and dataset augmentation for improving generalization and robustness.
%    \item \textbf{Saliency Guidance for Biometric PAD}: A discussion on how saliency-guided training for biometric PAD enhances model performance by aligning decision-making with human-salient regions.
%\end{enumerate}

\subsection{Neural Networks for Fingerprint PAD}

The Fingerprint Liveness Detection Competition (henceforth referred to as LivDet-Fingerprints), held every other year since 2009 \cite{livdet09}, illuminates the state of fingerprint presentation attack detection techniques throughout the years. Deep learning-based submissions first appeared in the 2015 edition of the competition \cite{livdet15}. In LivDet-Fingerprints 2023, the most recent edition of the competition \cite{livdet23}, every submission included some form of deep learning. The shift to deep learning can be attributed to the desire for generalization capabilities, since detecting unknown attack types is critical in LivDet competitions. Accordingly, the past five editions of the competition include novel attack types in their unseen testing sets \cite{livdet15, livdet17, livdet19, livdet21, livdet23}.

A common strategy in early neural network-based fingerprint PAD approaches was to analyze the fingerprint in patches, with each patch classification contributing as a vote toward the final decision. Decomposing training samples into multiple smaller patches, particularly from the corners and center, enhances model performance by increasing training diversity \cite{nogueira2016fingerprint}. Expanding datasets through patch reflection or normally-distributed random patch selection can also improve generalization \cite{park2016fingerprint}. Selecting patches based on the segmented foreground of a fingerprint can provide more relevant discriminatory information \cite{toosi2017cnn, zhang2019slim}, while minutiae-based patch selection has been shown to align well with key fingerprint features \cite{chugh2017fingerprint, chugh2018fingerprint}.

Fingerprint PAD research emphasizes strategies for handling and generalizing to unknown presentation attacks \cite{carta2025interpretability}. Dynamic adaptation of fingerprint liveness detectors has been described as a way to automatically learn and respond to previously unseen materials \cite{rattani2014automatic}. Expanding training data through style-transfer techniques has been demonstrated to improve model robustness against underrepresented attacks \cite{gajawada2019universal}. Further, synthetic fingerprint generation via style transfer can diversify training sets without requiring any spoof samples at all \cite{chugh2020fingerprint}.

\subsection{Saliency Guidance for Biometric PAD}
\label{subsec:SaliencyGuided}

Saliency-guided training primarily involves aligning model decision-making with regions of interest, denoted by saliency maps \cite{boyd2023cyborg}. Saliency guidance has been shown to enhance model generalization, making it well-suited for biometric PAD tasks \cite{Crum2024}.

This paper uses CYBORG loss $\mathcal{L}_\text{CYBORG}$ \cite{boyd2023cyborg} as a representative loss for saliency-guided training. It is a two-component loss composed of the classification component $\mathcal{L}_\text{classification}$ and the human saliency component $\mathcal{L}_\text{saliency-guidance}$, where:
\begin{equation}
\begin{split}
\mathcal{L}_\text{CYBORG} = &~\alpha*\mathcal{L}_\text{classification} \\
& + (1-\alpha)*\mathcal{L}_\text{saliency-guidance},\\
\end{split}
\end{equation}
\noindent and $\alpha$ weighs the two loss components. When $\alpha=1.0$, $\mathcal{L}_\text{CYBORG}$ becomes a standard cross-entropy loss, and when $\alpha=0.0$, $\mathcal{L}_\text{CYBORG}$ is only penalizing the model for lack of alignment between the model's and external saliencies, regardless of classification performance. In \cite{boyd2023cyborg}, saliency-guidance loss aligns Class Activation Mapping with human-sourced saliency provided for all training samples. By balancing human alignment with classification loss, models are more likely to learn discriminative features from human-salient regions. 

% (commonly cross-entropy \cite{boyd2023cyborg, Crum2023, crum2023teaching, Crum2024})

Saliency guidance has also been explored in a transformation-based approach. To preserve regions marked as important by provided saliency maps, non-salient unmarked regions are blurred \cite{Boyd2021}. This process aims to remove any distinguishing features from unannotated regions, effectively forcing model feature learning in areas highlighted by provided saliency maps. 

Saliency-guided training has demonstrated significant improvements in model generalization and classification accuracy when guided by both human- and algorithm-sourced saliency. Iris PAD models have benefited from guidance by human annotations \cite{Boyd2021}, AI student-generated saliency \cite{crum2023teaching}, autoencoder-generated saliency \cite{Crum2024}, and iris segmentation saliency \cite{Crum2024}. For synthetic face detection, the currently explored saliency types are human annotations \cite{boyd2023cyborg}, AI student-generated saliency \cite{crum2023teaching}, autoencoder-generated saliency \cite{Crum2024}, and face segmentation saliency \cite{Crum2024}. For both iris PAD and synthetic face detection, autoencoder-generated saliency, trained to predict human annotations on new samples, is the best explored saliency thus far. To maximize performance using saliency-guided training, the fidelity of conveyed information can be optimized through varying saliency granularities \cite{Crum2024}. %, such as: Features of Interest (FOI) provide high-fidelity (0–255) maps, Areas of Interest (AOI) represent binarized maps (0 or 255), and Boundaries of Interest (BOI) define the minimally enclosing rectangle around a saliency map's marked regions .

% Research Design
\section{Research Design}

Our research design consists of two main components: a human saliency collection study and an exploratory evaluation of saliency-guided fingerprint PAD models, trained with various saliency construction approaches aiming to guide the models towards features that improve generalization against unseen attack types.

\subsection{Justification of Data Selection}
We use the LivDet-Fingerprint datasets from 2015, 2017, 2019, and 2021 for both saliency collection and saliency-guided training \cite{livdet15, livdet17, livdet19, livdet21}. These datasets were chosen for three key reasons. First, they are the most recent and well-established benchmarks for evaluation of fingerprint PAD available to these authors. Second, their combined use provides a broader range of presentation attack instruments and capture devices, providing greater diversity than any single LivDet-Fingerprint competition alone. Finally, the LivDet-Fingerprint 2021 test set is the most challenging among our available datasets, as evidenced by its lower reported competitor scores compared to previous available LivDet-Fingerprint test sets. This increased difficulty, along with its inclusion of attack types not present in its training set or any prior LivDet dataset, makes it a strong choice for evaluating model performance.

\subsection{Construction of Limited Data Context}
\label{sec:LimitedData}

The training of biometric PAD models, including fingerprint PAD, is often limited by the availability of high-quality spoof samples \cite{karampidis2021comprehensive}. To simulate training conditions under a limited data availability context, we construct an 800-sample training dataset. 
This dataset size is selected to align with similar studies exploring low-data saliency strategies (\textit{e.g.} 765 in \cite{Boyd2021}). 
The limited data context is balanced by class, with 400 bonafide and 400 spoof samples. The spoof samples are further balanced over eight attack types at 50 samples each, sourced from the LivDet-Fingerprint 2015-2019 train and test datasets and the LivDet-Fingerprint 2021 train datasets: Ecoflex \cite{livdet15, livdet17, livdet19}, gelatine \cite{livdet15, livdet17, livdet19}, latex \cite{livdet15, livdet17, livdet19, livdet21}, liquid Ecoflex \cite{livdet15, livdet17, livdet19}, woodglue \cite{livdet15, livdet17, livdet19}, body double \cite{livdet17}, mix \cite{livdet19}, and RPRO Fast \cite{livdet21}. Bonafide and spoof samples are further balanced by the capture sensors represented by each subclass (note that not all attack types are represented by all scanners). 

\subsection{Human-Annotative Saliency Collection}
\label{sec:Collection}

Biometric saliency guidance research is based on human-annotative saliency, which, to our knowledge, has not been previously collected for the fingerprint PAD task. To support our exploration and future work in this domain, we conducted an annotation collection study. An example image annotated by a subject to indicate features supporting their decision is illustrated in Fig. \ref{fig:teaser}.

% The unit of analysis is each fingerprint scan along with its corresponding annotations.
% \footnote{details about the acquisition site and population has been removed to preserve paper's anonymity}
We recruited 50 participants from the University of Notre Dame using a convenience sampling technique due to accessibility, scheduling, and other practical constraints.
We collected annotations over the limited 800-sample dataset described in Sec. \ref{sec:LimitedData}.
Each participant annotated 32 fingerprint scans, 16 bonafide and 16 spoof samples, with spoof samples spanning eight attack types (two images per type).
Participants used an image annotation software specially designed for this study to (a) predict each sample’s class (genuine or fake), (b) annotate salient regions supporting their decision, and (c) optionally provide textual description to further support their classification decisions ({\it e.g.}, in cases when it would be difficult to support subject's decision by annotating the image, such as perfect symmetry of the fingerprint image, or unnatural contrast, or too dark image). 
Participants were not instructed to annotate any specific fingerprint features (\textit{e.g.} whorls, ridges), as intuitively-annotated regions mitigate feature bias, promote straightforward collection reproducibility, and better align with the philosophy of saliency guidance.
The created image annotation software also recorded start and end time stamps of each stroke. This allows for fine-grained analysis of annotation time and provides the order of features annotated by each subject on each sample. 
%The metrics for key variables, annotated regions on fingerprint scans, are nominal.
Confounding factors in this study include the level of biometric expertise or fingerprint familiarity among annotators. However, we consider this factor to be of limited concern, as prior saliency collection studies have demonstrated success with non-expert annotators \cite{Boyd2021, boyd2023cyborg}.
%The target population for this data is biometrics researchers and those studying saliency-guided or human-inspired learning techniques.

\subsection{Construction of Saliency Types}
\label{sec:SaliencyTypes}

\begin{figure*}[!htb]
    \centering
    \resizebox{\textwidth * 0.8}{!}{%
    \includegraphics[width=\linewidth, trim = 2cm 4cm 2cm 4cm]{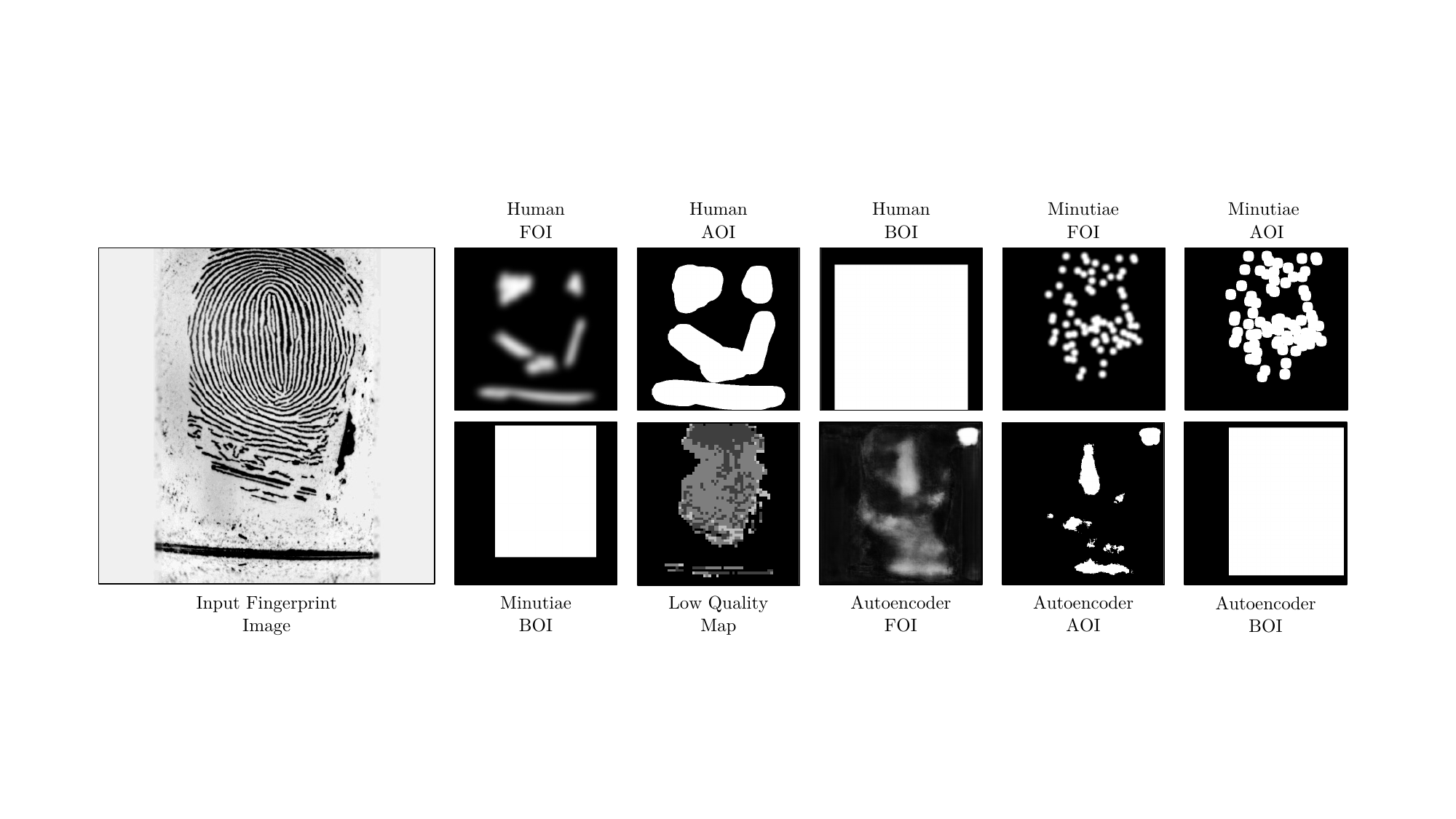}
    }
    \caption{Illustration of all saliency types explored in our saliency-guided training experiments for an example fingerprint image. FOI, AOI, and BOI refer to Features-, Area-, and Boundary of Interest saliency granularities, respectively.}
    \label{tab:saliency-types}
\end{figure*}

We construct several saliency types for our training experiments, including human-annotated saliency and algorithmically-sourced alternatives, which we term \textit{pseudosaliency}. All types of saliency are displayed in Fig. \ref{tab:saliency-types}.

\subsubsection{Human-Annotative Saliency}
\textbf{Human-annotative saliency} maps are derived from human annotations collected in the study described in Sec. \ref{sec:Collection}. Following established methodology \cite{boyd2023cyborg}, each map is created by averaging the saliency annotations from two independent annotators. We derive three levels of saliency granularity using techniques described in \cite{Crum2024}: 

\begin{itemize}
    \item Features of Interest (FOI), the original continuous-valued human-annotative saliency map,
    \item Area of Interest (AOI), a binarized version of the FOI, and 
    \item Boundary of Interest (BOI), the minimally enclosing rectangle around the AOI.
\end{itemize}

\subsubsection{Pseudosaliency Maps}
We further produce several types of \textit{pseudosaliency}, which we define as saliency that conveys human-interpretable, domain-specific information, but is not directly drawn or labeled by humans.

\textbf{Minutiae-based pseudosaliency} maps are constructed from minutiae points extracted using the Neurotechnology VeriFinger 13.1 SDK \cite{verifinger}. Each detected minutia is drawn as a circular region with a radial falloff, using a radius of 10 pixels. FOI, AOI, and BOI granularities are derived using the same transformation process as in human-annotative saliency. The intuition behind this pseudosaliency type is that regions considered important for the fingerprint recognition task may translate in importance to fingerprint PAD.

\textbf{Low-quality pseudosaliency} maps indicate fingerprint regions with poor pattern fidelity. These are constructed using NBIS-generated quality and low-contrast maps \cite{nbis}. Specifically, the quality map is inverted to emphasize low-quality areas, the low-contrast map is also inverted to segment the fingerprint pattern, and the final saliency map is obtained by masking the inverted quality map with the inverted low-contrast map. This creation process preserves low-quality regions that lie on the fingerprint pattern. The intuition behind this pseudosaliency type is that low-quality regions in the image may reflect poor imitations of genuine fingerprint features, suggesting a possible spoof.

\textbf{Autoencoder-generated pseudosaliency} maps are generated to mimic human annotations on unannotated fingerprint samples. We employ a DenseNet-161-based U-Net architecture \cite{autoencoder}, trained using human-annotated samples, following prior autoencoder-based strategy \cite{Crum2024}. The resulting maps are treated as FOI-level saliency. AOI is obtained by thresholding at 0.5 to account for autoencoder artifacts, and BOI is computed by minimally enclosing AOI. 

\subsection{Training of PAD Models and Metrics Used}

% The unit of analysis is each fingerprint PAD model configuration, evaluated across multiple training contexts. 
To select model training configurations, we used maximum variation purposive sampling to account for variations in saliency types, loss balancing weights, model architectures, and data contexts.
The data collection technique is indirect, involving analysis of usage logs: each model’s predictions on a shared testing set (the LivDet-Fingerprint 2021 testing set).
The primary metrics of analysis are each model’s testing classification accuracy at the validation Equal Error Rate-equivalent threshold and Area Under the ROC curve (AUC).
Key confounding factors in evaluating machine learning models include class imbalance and overfitting. We mitigate these by balancing training data (ensuring equal distribution of bonafide/spoof samples, attack types, and scanner types) and testing on unseen data that includes novel attack types.

% Experimental Scenarios
\section{Experimental Training Scenarios}
\label{sec:Scenarios}

To assess the impact of saliency-guided training on fingerprint PAD, we designed a series of training scenarios that explore different model configurations under varying data availability constraints and saliency integration methods. These experiments compare saliency-guided configurations against both baseline techniques and each other, given the sensitivity of saliency guidance to implementation choices.

All models are evaluated on LivDet-Fingerprint 2021 testing set, which contains novel attack types, was never used in training, and enables direct comparison with algorithms submitted to the competition. Further, we train each configuration on three convolutional neural network (CNN) architectures: ResNet50 \cite{resnet}, DenseNet-121 \cite{densenet}, and Inception-V3 \cite{inception}. %, selected for their prior use in saliency-guided PAD research \cite{boyd2023cyborg, Crum2024}. 
Each configuration undergoes three independent training runs to assess the variability of point error estimates (accuracy and AUC). All training and testing samples are center-cropped and resized to $224 \times 224$ pixel resolution.

\subsection{Scenario S1: Large Data Baseline}
To establish a benchmark under large data availability conditions, we train models on a comprehensive dataset containing all available LivDet-Fingerprint samples (certainly excluding the LivDet-Fingerprint 2021 test set), totaling 59,792 fingerprint images. Validation is randomly selected at 20\%, and the models are trained using cross-entropy loss, hence without any saliency guidance.

\subsection{Scenario S2: Limited Data Baseline}
To simulate limited-data conditions, typical for PAD problems, we train models on a reduced 800-sample dataset (the same dataset used for human-annotative saliency collection) using cross-entropy loss. Again, no saliency information is used guide the models during training. This scenario highlights the challenges of learning from limited spoof samples and serves as a baseline for saliency-guided methods.

\subsection{Scenario S3: Limited Data Availability with Loss-Based Saliency-Guidance}
Using the same 800-sample dataset as in S2, but now with saliency information, we implement saliency guidance via the CYBORG loss function \cite{boyd2023cyborg}, which aligns model's Class Activation Maps with external saliency during training to encourage learning in relevant regions. To explore the effect of saliency weighting, we experiment with different values of the CYBORG loss weighting parameter $\alpha$, which determines the proportion of cross-entropy loss versus saliency alignment loss. Specifically, we test $\alpha$=\{0.1,0.3,0.5,0.7,0.9\}, covering a wide range from heavily CAM-alignment training to predominantly cross-entropy-based learning. We evaluate on the following saliency types, defined in Sec. \ref{sec:SaliencyTypes}: human-annotative saliency (FOI, AOI, BOI), minutiae-based pseudosaliency (FOI, AOI, BOI), and low-quality pseudosaliency.

% We evaluate the following saliency sources and their respective fidelity granularities \cite{Crum2024}:

% \begin{itemize}
%     \item \textbf{Human-Annotative Saliency (FOI, AOI, BOI):} Saliency maps generated from human annotations, with each sample composed by evenly combining two annotators, following prior methodologies \cite{boyd2023cyborg}. FOI represents full-fidelity maps, AOI is a binarized version, and BOI is the minimally enclosing rectangle around AOI.
%     \item \textbf{Minutiae-Based Saliency (FOI, AOI, BOI):} Saliency maps based on minutiae points detected by Neurotechnology VeriFinger 13.1 SDK, represented as radially fading circles (radius = 10). FOI, AOI, and BOI are derived using the same transformation process as in human-annotative saliency.
%     \item \textbf{Quality-Based Saliency (FOI only):} Saliency maps derived from NBIS-generated quality and low-contrast segmentation maps \cite{nbis}, emphasizing low-quality fingerprint regions. The quality map is inverted, masked by the low-contrast map, and normalized to communicate low-quality regions within the fingerprint texture.
% \end{itemize}

\begin{table*}
\centering
\small
\setlength{\tabcolsep}{6pt}
\renewcommand{\arraystretch}{1.2}

\resizebox{\textwidth}{!}{%
\begin{tabular}{@{}lcccc|cccccc@{}}
\toprule
\textbf{Saliency} & \textbf{Granularity} & \textbf{Network} & \textbf{Loss} & \textbf{$\alpha$} & \textbf{Accuracy} & \textbf{Bonafide Acc.} & \textbf{Spoof Acc.} & \textbf{AUC} \\
\midrule

% S1
\rowcolor{gray!25}
\multicolumn{9}{l}{\textit{\textbf{S1}: Large Data Baseline}} \\

- & - & ResNet & Cross Entropy & - & 0.911±0.001 & 0.995±0.000 & 0.821±0.003 & 0.989±0.000 \\ 

\midrule

% S2
\rowcolor{gray!25}
\multicolumn{9}{l}{\textit{\textbf{S2}: Limited Data Baseline}} \\

- & - & Inception & Cross Entropy & - & 0.862±0.010 & 0.948±0.009 & 0.769±0.014 & 0.946±0.007 \\ 

\midrule

% S3
\rowcolor{gray!25}
\multicolumn{9}{l}{\textit{\textbf{S3}: Limited Data Availability with Loss-Based Saliency Guidance}} \\

Human & AOI & Inception & CYBORG & 0.9 & 0.879±0.006 & \underline{0.966±0.002} & 0.786±0.012 & 0.959±0.002 \\
Minutiae & FOI & Inception & CYBORG & 0.7 & \underline{0.885±0.004} & 0.952±0.006 & \underline{0.812±0.014} & \underline{0.961±0.004} \\
Low Quality & - & Inception & CYBORG & 0.5 & 0.876±0.011 & 0.953±0.014 & 0.795±0.019 & 0.951±0.005 \\

\midrule

% S4
\rowcolor{gray!25}
\multicolumn{9}{l}{\textit{\textbf{S4}: Limited Data Availability with Blur-Based Saliency Guidance}} \\

- & - & Inception & Cross Entropy & - & 0.794±0.018 & 0.869±0.019 & 0.714±0.057 & 0.878±0.013 \\
Human & BOI & Inception & Cross Entropy & - & \underline{0.873±0.006} & \underline{0.958±0.007} & 0.781±0.015 & \underline{0.953±0.004} \\
Minutiae & AOI & Inception & Cross Entropy & - & 0.868±0.008 & 0.907±0.008 & \underline{0.825±0.020} & 0.942±0.006 \\
Low Quality & - & Inception & Cross Entropy & - & 0.828±0.018 & 0.843±0.020 & 0.813±0.019 & 0.906±0.016 \\

\midrule

% S5
\rowcolor{gray!25}
\multicolumn{9}{l}{\textit{\textbf{S5}: Large-Data Availability with Pseudosaliency Guidance}} \\

Autoencoder & AOI & Inception & CYBORG & 0.9 & 0.916±0.011 & 0.994±0.001 & 0.831±0.024 & \textbf{\underline{0.990±0.001}} \\
Minutiae & AOI & DenseNet & CYBORG & 0.9 & \textbf{\underline{0.938±0.004}} & 0.964±0.001 & \textbf{\underline{0.910±0.009}} & 0.988±0.001\\
Low Quality & - & Inception & CYBORG & 0.9 & 0.906±0.011 & \textbf{\underline{0.996±0.001}} & 0.808±0.023 & 0.808±0.023 \\

\bottomrule
\end{tabular}%
}

\vspace{0.5em}

% \caption{\textbf{INCOMPLETE}}
% \label{tab:large-summary}

\caption{Summary of best-accuracy-achieving configurations across five training scenarios. Each row describes a configuration as well as its key performance metrics. 
For each metric, the highest inter-scenario score is \textbf{bolded} and the highest intra-scenario score is \underline{underlined}. Full scenario results are available in the supplementary materials.}

\label{tab:scenario-summary}

\end{table*}

\subsection{Scenario S4: Limited Data Availability with Blur-Based Saliency Guidance}

Following the same 800-sample dataset structure, we apply saliency guidance with cross-entropy loss through a transformation-based strategy \cite{Boyd2021}, where non-salient regions are Gaussian blurred (radii = \{2, 4, 6, 8, 10, 12, 14, 16\}) to ``remove'' (by blurring) irrelevant and emphasize important features, increasing the effective training set size eightfold. To mitigate edge artifacts between blurred and unblurred regions, the saliency masks are smoothed with a Gaussian blur (radius = 5). We explore the same saliency types as those used in scenario S3.

As a control, we train models on fully blurred images at the same radii, alongside the original unblurred images, expanding the training set ninefold. This ensures that observed performance differences are from retained salient regions rather than dataset generic (non-saliency-related) augmentations.

\subsection{Scenario S5: Large Data Availability with Loss-Based Pseudosaliency Guidance} 

Using the autoencoder-generated pseudosaliency process described in Sec. \ref{sec:SaliencyTypes}, we generate FOI, AOI, and BOI pseudosaliency for the large data (omitting the 800 human-annotated samples used to train the autoencoder). We further produce minutiae-based (FOI, AOI, and BOI) and quality-based pseudosaliency for the expanded data. This scenario evaluates whether non-human pseudosaliency can effectively scale saliency-guided training to larger datasets.

% Results
\section{Results}

This section presents the results of the five training scenarios by directly addressing the research questions listed in Sec. \ref{sec:Introduction}. A summary of the top-performing configurations for each scenario is provided in Tab. \ref{tab:scenario-summary}.

\subsection*{Addressing RQ1: Does saliency-guided training with human-annotative saliency improve generalization and classification accuracy?}

Model training with human saliency is explored in scenarios S3 and S4, and its best configurations are described in Tab. \ref{tab:scenario-summary}. In both loss- and blur-based saliency guidance, human-guided configurations outperform their comparative limited-data baselines (scenarios S2 and S4).

Among human-saliency configurations with limited-data loss-based guidance (scenario S3), the configuration achieving the highest accuracy is configured with the Area of Interest saliency granularity, the Inception architecture, and a CYBORG $\alpha$ weight of 0.9. Its achieved accuracy of 0.879±0.006 and AUC of 0.959±0.002 outperform the best baseline configuration (scenario S2) by 0.017 and 0.013, respectively.

Continually, the best human saliency blur-guided model (scenario S4), using the Boundary of Interest granularity and the Inception architecture, achieves a classification accuracy of 0.873±0.006 and an AUC of 0.953±0.004, surpassing the blurring baseline by 0.079 and 0.075, respectively.

These results demonstrate that \textbf{saliency-guided training with human-annotative saliency improves generalization and classification accuracy for fingerprint PAD.} 

\begin{table*}[!htb]
\centering
% \scriptsize

%\resizebox{\columnwidth}{!}{%
\begin{tabular}{@{}l|c|c|c@{}}
\toprule
\textbf{Domain} & \textbf{Baseline} & \textbf{Saliency-Guided Training} & \textbf{Normalized Gain} \\
\midrule
Iris PAD \cite{Crum2024} & 0.893±0.019 & 0.962±0.005 & +64.5\% \\
Synthetic Face Detection \cite{Crum2024} & 0.572±0.047 & 0.643±0.033 & +16.6\% \\
\midrule
\textbf{Fingerprint PAD (limited)} & 0.946±0.007 & 0.961±0.004 & +27.8\% \\
\textbf{Fingerprint PAD (large)} & 0.990±0.001 & 0.991±0.001 & +10.0\% \\
\bottomrule
\end{tabular}
%}

\vspace{1em}

\caption{Normalized generalization gain of the highest-AUC-scoring saliency-guided models across four biometric attack detection tasks.}

\label{tab:norm-gain-simple}

\end{table*}

\subsection*{Addressing RQ2: Are algorithmically-sourced pseudosaliency types effective for saliency-guided training of fingerprint PAD?}
\label{subsec:RQ2}

Table \ref{tab:scenario-summary} outlines the best training configurations for pseudosaliency-guided models, as explored in scenarios S3, S4, and S5. In the limited data context, all explored pseudosaliency types, when optimally configured, are capable of outperforming baseline models (scenarios S2 and S4). In the large-data context, two of the three proposed pseudosaliency types exceed baseline performance (scenario S1).

For loss-based pseudosaliency guidance in the limited-data context (scenario S3), minutiae-based Features of Interest saliency type achieves the highest classification accuracy (0.885±0.004) and AUC (0.961±0.004), surpassing the baseline metrics (scenario S2) by 0.023 in accuracy, and by 0.015 in AUC. This is the best performance exhibited in the limited data context, outperforming human-annotative saliency.

For blur-based guidance (scenario S4), the best pseudosaliency-guided model uses the minutiae-based Area of Interest saliency type, achieving a classification accuracy of 0.868±0.008 and an AUC of 0.942±0.006, outperforming the blurring baseline but not the best human saliency configuration.

The generative nature of pseudosaliency enables saliency-guided training to scale effectively in larger data availability contexts, as explored in scenario S5. The best configuration in this scenario employs minutiae-based Area of Interest pseudosaliency, achieving an accuracy of 0.938±0.004. This result outperforms the large-data baseline (scenario S1) by 0.027. When compared to the reported LivDet-Fingerprint 2021 rankings \cite{livdet21}, {\bf this performance would have earned or tied for first place}, accounting for our reported standard deviation (which was not included in the LivDet-Fingerprint 2021 results). This configuration raises spoof detection accuracy by 8.9\%, demonstrating the generalization strength of saliency-guided training. 

Thus, we can affirm that \textbf{algorithmically-sourced pseudosaliency is effective for saliency-guided fingerprint PAD training.} Moreover, it is capable of exceeding human-annotative saliency performance as well as scaling to large data contexts, even if not directly designed for the fingerprint PAD task. 

\subsection*{Addressing RQ3: How does saliency-guided training affect fingerprint PAD generalization compared to other biometric attack detection tasks?}

As in answers to the research questions RQ1 and RQ2, saliency-guided training enhances both generalization and classification performance for fingerprint PAD, further affirming its applicability to biometric attack detection tasks. However, comparing performance across domains requires understanding baseline differences. Considering achieved AUC, fingerprint PAD baselines already perform relatively well compared to iris presentation attack detection and synthetic face detection \cite{Boyd2021, boyd2023cyborg, Crum2024}, meaning there is less room for improvement.

To evaluate the impact of saliency-guided training on model generalization while accounting for baseline performance, we compute normalized gain $g$, defined for a metric $M$ as:

\begin{equation}
g_\text{ normalized} = \frac{M_{\text{ saliency-guided}} - M_{\text{ baseline}}}{100\% - M_{\text{ baseline}}}
\end{equation}

Tab. \ref{tab:norm-gain-simple} summarizes normalized AUC performance gains for iris PAD, synthetic face detection, and fingerprint PAD in both large and limited data contexts. The generalization gains observed for fingerprint PAD are most comparable to those in synthetic face detection but less than improvements in iris PAD, as reported in \cite{Crum2024}. Thus, we conclude that \textbf{saliency-guided training positively impacts fingerprint PAD generalization to a similar or lesser extent than other biometric domains.}

% Discussion
\section{Discussion}

Through our explored scenarios and answered research questions, we provide a foundational understanding of saliency-guided training for fingerprint PAD. In this section, we will highlight key observations and broader takeaways from our findings.

\subsection{Improvements to Bonafide vs. Spoof Accuracy}

A closer look at model performance metrics in Tab. \ref{tab:scenario-summary} reveals that saliency-guided training primarily raises spoof detection accuracy, with bonafide accuracy usually affected to a lesser degree. In the limited-data context, the best configuration (Inception, CYBORG, $\alpha$=0.7, Minutiae FOI) improves spoof accuracy by 4.3\%, while bonafide accuracy increases by only 0.4\%. Similarly, in the large-data context, the best configuration (DenseNet, CYBORG, $\alpha$=0.9, Minutiae AOI) gains 8.9\% in spoof accuracy but loses 3.1\% in bonafide accuracy. 

\subsection{CNN Architecture Sensitivity}

We observe patterns among high-performing configurations that suggest broader trends in saliency-guided training for fingerprint PAD. Notably, the Inception-V3 architecture consistently performs well, appearing in all but two top-accuracy-achieving models across best explored configurations described in Tab. \ref{tab:scenario-summary}. This contrasts with other explored biometric attack detection tasks: DenseNet performs best on average for iris PAD, while ResNet and DenseNet alternate depending on loss in synthetic face detection \cite{Crum2024}. These results indicate that there is no optimal architecture for saliency-guided training across biometric PAD domains. This further describes how this training paradigm is highly sensitive to configuration, including saliency type and granularity, loss weighting, and size of training data.

\subsection{Blurring and BOI Saliency Granularity}

In blur-based saliency guidance, the Boundary of Interest (BOI) saliency granularity consistently performs well. It is the best granularity for human saliency and second best among explored pseudosaliency types. BOI saliency identifies a minimal enclosing rectangle around all regions of interest, which typically encompasses most of the fingerprint area. Consequently, when non-salient regions are blurred, the fingerprint pattern remains largely unaffected. Combined with eightfold data augmentation through multiple blur radii, the model encounters the nearly unblurred pattern repeatedly. This inadvertently increases exposure to samples during training, similar to training with standard cross entropy for many additional epochs. While BOI saliency achieves high scores on key metrics, this effectiveness is likely due to unintended augmentation behavior rather than true saliency-based guidance, which impacts how blur-based saliency guidance results may be interpreted.

\subsection{Effectiveness of Pseudosaliency Types}

Software-generated pseudosaliency, specifically minutiae- and low-quality-based types, proved effective for fingerprint PAD guidance. Models trained with these sources matched or exceeded the performance of those trained with human-annotative saliency (scenario S3). This contrasts with prior findings in other biometric domains \cite{Crum2024}, where image segmentation-based saliency for iris and face images resulted in sub-baseline performance for both iris PAD and synthetic face detection. Autoencoder-generated pseudosaliency also proved effective for fingerprint PAD (scenario S5), outperforming the baseline in the large-data scenario (scenario S1). This is consistent with previous findings where autoencoder-based saliency surpassed human annotations in both iris and face domains \cite{Crum2024}.

From these results, we hypothesize that effective pseudosaliency must communicate domain-specific knowledge. Minutiae-based saliency highlights distinguishing fingerprint features used in recognition. Low-quality maps carry meaningful information about a fingerprint's fidelity. Autoencoder-generated saliency is inspired by human annotation training data collected directly for the fingerprint PAD task. In contrast, image segmentation models simply describe object boundaries without offering domain-specific insights. Our findings suggest that algorithmically-sourced pseudosaliency maps, when grounded in domain-specific knowledge, offer a scalable and cost-effective alternative to human annotations, particularly when human annotations are difficult or expensive to obtain.

\subsection{Scalability and Future Directions of Saliency-Guided Training}

Scenario S5 offers a promising insight into the potential scalability of saliency-guided training. Prior work has shown that while saliency-guided training can provide gains in limited-data contexts, it typically fails to surpass performance achieved with large-scale training alone \cite{Boyd2021}. Our results in limited-data scenarios (S2, S3, S4) reflect this trend. However, by incorporating generative pseudosaliency in a large-data context (scenario S5), we observe accuracy and generalization improvements over the large-data baseline (scenario S1) across several configurations.

The results of scenario S5 represent a meaningful step forward: it suggests that generative saliency approaches may help scale saliency-guided training to previously inapplicable large datasets. These findings point toward future directions where pseudosaliency guidance at-scale could positively complement traditional training.

% Conclusion
\section{Conclusion}

This paper explores the intersection of saliency-guided training and fingerprint presentation attack detection, examining how saliency types, guidance strategies, and data availability contexts influence model behavior. Across five experimental scenarios, for the first time we evaluate the impact of human-annotative saliency and algorithmically-sourced pseudosaliency in guiding the training of fingerprint PAD models. Our findings reveal meaningful benefits in classification and generalization performance, though to a lesser extent than in other biometric domains. We highlight the configuration-sensitive nature of saliency-guided training and identify promising directions for scalable, domain-aware saliency generation. 

To support future research and facilitate full replicability of this study, we release both human-annotated saliency and algorithmically-sourced pseudosaliency data, along with source codes and model weights. Overall, this work contributes to a deeper understanding of how saliency-guided training strategies behave within the fingerprint PAD domain.

\paragraph{Acknowledgment} This material is based upon work supported by the U.S. National Science Foundation under grant No. 2237880. Any opinions, findings, conclusions, or recommendations expressed in this material are those of the author(s) and do not necessarily reflect the views of the U.S. National Science Foundation.

{\small
\bibliographystyle{ieee}
\bibliography{egbib}
}

% Supplementary Materials - Remove for Submission PDF
% \newpage
\onecolumn

\setcounter{section}{0}
\renewcommand*{\thesection}{Supp-Mat-\the\value{section}}

\appendix
% \section*{Saliency-Guided Training for Fingerprint Presentation Attack Detection \newline Samuel Webster \& Adam Czajka, University of Notre Dame \newline
% Supplementary Materials}

\section*{Appendix}

\noindent The Appendix contains the complete experimental results for all five training scenarios (S1–S5) explored in the main paper. Each table reports fingerprint PAD model performance across different saliency types, saliency granularities, and network architectures. All configurations are averaged over three independent runs. All testing accuracy metrics are computed using the Equal Error Rate threshold from each model's validation set predictions. We define all listed columns as:

\begin{itemize}
    \item \textbf{Network}: The CNN network architecture used from: ResNet50, DenseNet-121, and Inception-V3.
    \item \textbf{Alpha ($\alpha$)}: Refers to the $\alpha$ parameter of $\mathcal{L}_\text{CYBORG}$ loss, which weighs the $\mathcal{L}_\text{classification}$ and $\mathcal{L}_\text{saliency-guidance}$ parameters. At 0.0, loss is saliency guidance only, and at 1.0, loss is cross entropy only.
    \item \textbf{AUC}: Area under the receiver operating characteristic (ROC) curve.
    \item \textbf{Accuracy}: Overall classification accuracy on the LivDet-Fingerprint 2021 test set.
    \item \textbf{Bonafide Accuracy}: Classification accuracy on bonafide (live, real) fingerprint samples from the LivDet-Fingerprint 2021 test set.
    \item \textbf{Spoof Accuracy}: Classification accuracy on spoof (fake) fingerprint samples from the LivDet-Fingerprint 2021 test set.
    \item \textbf{Placement}: A hypothetical ranking in the LivDet-Fingerprint 2021 competition based on model-achieved Accuracy and reported competitor-achieved accuracies. The placement range is computed over $\mu_\text{ accuracy}\pm\sigma_\text{ accuracy}$.  
    \item \textbf{d'}: Measures how well models can distinguish between the live and spoof classes. Higher values indicate greater separability.
    \item \textbf{FNR @ FPR = 1\%}: The False Negative Rate at the threshold where the False Positive Rate is 1\%. Lower value is better.
    \item \textbf{$\text{\textit{Attack Type}}^\text{\textit{U/K}}$ Accuracy}: Classification accuracy across the various spoof (fake) types in the LivDet-Fingerprint 2021 test set. A superscript $^\text{U}$ indicates an unknown attack type (no samples of this attack type seen during training) while a superscript $^\text{K}$ indicates a known attack type (samples of this attack type seen during training).
\end{itemize}

\section{Scenario S1, Full Results}

\FloatBarrier

\begin{table}[H]
\centering

\label{tab:appendix-scenario1-results}

\resizebox{\textwidth}{!}{%
\begin{tabular}{@{}l|c|c|c|c|c|c|c|c|c|c|c|c@{}}
\toprule
\textbf{Network} & \textbf{AUC} & \textbf{Accuracy} & \makecell{\textbf{Bonafide} \\ \textbf{Accuracy}} & \makecell{\textbf{Spoof} \\ \textbf{Accuracy}} & \textbf{Placement} & \textbf{d'} & \makecell{\textbf{FNR @} \\ \textbf{FPR = 1\%}} & \makecell{\textbf{$\text{GLS20}^\text{U}$} \\ \textbf{Accuracy}} & \makecell{\textbf{$\text{VBN}^\text{U}$} \\ \textbf{Accuracy}} & \makecell{\textbf{$\text{RFastPro}^\text{U}$} \\ \textbf{Accuracy}} & \makecell{\textbf{$\text{Wood Glue}^\text{K}$} \\ \textbf{Accuracy}} & \makecell{\textbf{$\text{Body Double}^\text{K}$} \\ \textbf{Accuracy}} \\ \midrule

\makecell{ResNet \\ DenseNet \\ Inception} & \makecell{0.989±0.000 \\ 0.990±0.001 \\ 0.990±0.010} & \makecell{0.911±0.001 \\ 0.905±0.003 \\ 0.911±0.008} & \makecell{0.995±0.000 \\ 0.995±0.001 \\ 0.996±0.001} &
\makecell{0.821±0.003 \\ 0.808±0.006 \\ 0.819±0.018} & \makecell{4 \\ 4 \\ 3-4} &  \makecell{3.261±0.020 \\ 3.317±0.045 \\ 3.273±0.072} & \makecell{0.133±0.004 \\ 0.146±0.010 \\ 0.137±0.006} & \makecell{0.747±0.007 \\ 0.725±0.008 \\ 0.731±0.035} & \makecell{0.893±0.004 \\ 0.898±0.001 \\ 0.914±0.017} & \makecell{0.780±0.003 \\ 0.751±0.004 \\ 0.757±0.045} & \makecell{0.890±0.020 \\ 0.899±0.023 \\ 0.904±0.017} & \makecell{0.868±0.004 \\ 0.857±0.014 \\ 0.884±0.009}\\

\bottomrule
\end{tabular}%
}

\vspace{0.5em} 

\caption{\textbf{Results for Scenario 1.} Models trained using cross entropy in the large-data-availability context. Averaged over three runs.}

\end{table}

\section{Scenario S2, Full Results}

\begin{table}[H]
\centering

\label{tab:appendix-scenario2-results}

\resizebox{\textwidth}{!}{%
\begin{tabular}{@{}l|c|c|c|c|c|c|c|c|c|c|c|c@{}}
\toprule
\textbf{Network} & \textbf{AUC} & \textbf{Accuracy} & \makecell{\textbf{Bonafide} \\ \textbf{Accuracy}} & \makecell{\textbf{Spoof} \\ \textbf{Accuracy}} & \textbf{Placement} & \textbf{d'} & \makecell{\textbf{FNR @} \\ \textbf{FPR = 1\%}} & \makecell{\textbf{$\text{GLS20}^\text{U}$} \\ \textbf{Accuracy}} & \makecell{\textbf{$\text{VBN}^\text{U}$} \\ \textbf{Accuracy}} & \makecell{\textbf{$\text{RFastPro}^\text{U}$} \\ \textbf{Accuracy}} & \makecell{\textbf{$\text{Wood Glue}^\text{K}$} \\ \textbf{Accuracy}} & \makecell{\textbf{$\text{Body Double}^\text{K}$} \\ \textbf{Accuracy}} \\ \midrule

\makecell{ResNet \\ DenseNet \\ Inception} & \makecell{0.930±0.003 \\ 0.918±0.016 \\ 0.946±0.007} &  \makecell{0.847±0.003 \\ 0.839±0.012 \\ 0.862±0.010} & \makecell{0.928±0.007 \\ 0.944±0.004 \\ 0.948±0.009} &
\makecell{0.760±0.014 \\ 0.725±0.025 \\ 0.769±0.014} & \makecell{5 \\ 5 \\ 5} &  \makecell{2.093±0.027 \\ 1.978±0.161 \\ 2.283±0.094} & \makecell{0.525±0.042 \\ 0.518±0.063 \\ 0.474±0.027} & \makecell{0.747±0.007 \\ 0.725±0.008 \\ 0.731±0.035} & \makecell{0.893±0.004 \\ 0.898±0.001 \\ 0.914±0.017} & \makecell{0.780±0.003 \\ 0.751±0.004 \\ 0.757±0.045} & \makecell{0.890±0.020 \\ 0.899±0.023 \\ 0.904±0.017} & \makecell{0.868±0.004 \\ 0.857±0.014 \\ 0.884±0.009}\\

\bottomrule
\end{tabular}%
}

\vspace{0.5em} 

\caption{\textbf{Results for Scenario 2.} Models trained using cross entropy in the limited-data context. Averaged over three runs.}

\end{table}

\section{Scenario S3, Full Results}

\begin{table}[H]
\centering

\label{tab:appendix-scenario3-human-foi-results}

\resizebox{\textwidth}{!}{%
\begin{tabular}{@{}l|c|c|c|c|c|c|c|c|c|c|c|c|c@{}}
\toprule
\textbf{Network} & \textbf{Alpha ($\alpha$)} & \textbf{AUC} & \textbf{Accuracy} & \makecell{\textbf{Bonafide} \\ \textbf{Accuracy}} & \makecell{\textbf{Spoof} \\ \textbf{Accuracy}} & \textbf{Placement} & \textbf{d'} & \makecell{\textbf{FNR @} \\ \textbf{FPR = 1\%}} & \makecell{\textbf{$\text{GLS20}^\text{U}$} \\ \textbf{Accuracy}} & \makecell{\textbf{$\text{VBN}^\text{U}$} \\ \textbf{Accuracy}} & \makecell{\textbf{$\text{RFastPro}^\text{U}$} \\ \textbf{Accuracy}} & \makecell{\textbf{$\text{Wood Glue}^\text{K}$} \\ \textbf{Accuracy}} & \makecell{\textbf{$\text{Body Double}^\text{K}$} \\ \textbf{Accuracy}} \\ \midrule

\makecell{ResNet \\ DenseNet \\ Inception} & 0.9 & \makecell{0.936±0.006 \\ 0.944±0.005 \\ 0.953±0.003} & \makecell{0.860±0.012 \\ 0.861±0.009 \\ 0.872±0.003} & \makecell{0.934±0.011 \\ 0.952±0.008 \\ 0.957±0.003} & \makecell{0.779±0.035 \\ 0.763±0.012 \\ 0.780±0.004} & \makecell{4-5 \\ 5 \\ 4-5} & \makecell{2.153±0.068 \\ 2.254±0.057 \\ 2.365±0.037} & \makecell{0.510±0.043 \\ 0.448±0.039 \\ 0.446±0.008} & \makecell{0.696±0.058 \\ 0.679±0.025 \\ 0.705±0.013} & \makecell{0.872±0.015 \\ 0.825±0.052 \\ 0.863±0.026} & \makecell{0.733±0.056 \\ 0.719±0.015 \\ 0.736±0.023} & \makecell{0.845±0.030 \\ 0.825±0.025 \\ 0.831±0.068} & \makecell{0.836±0.007 \\ 0.842±0.009 \\ 0.841±0.010} \\
\bottomrule
\end{tabular}%
}

\vspace{0.5em} 

\caption{\textbf{Results for Scenario S3 using human-annotated FOI saliency}. Models trained using CYBORG loss in the limited-data context. Averaged over three runs.}

\end{table}

\begin{table}[H]
\centering

\label{tab:appendix-scenario3-human-aoi-results}

\resizebox{\textwidth}{!}{%
\begin{tabular}{@{}l|c|c|c|c|c|c|c|c|c|c|c|c|c@{}}
\toprule
\textbf{Network} & \textbf{Alpha ($\alpha$)} & \textbf{AUC} & \textbf{Accuracy} & \makecell{\textbf{Bonafide} \\ \textbf{Accuracy}} & \makecell{\textbf{Spoof} \\ \textbf{Accuracy}} & \textbf{Placement} & \textbf{d'} & \makecell{\textbf{FNR @} \\ \textbf{FPR = 1\%}} & \makecell{\textbf{$\text{GLS20}^\text{U}$} \\ \textbf{Accuracy}} & \makecell{\textbf{$\text{VBN}^\text{U}$} \\ \textbf{Accuracy}} & \makecell{\textbf{$\text{RFastPro}^\text{U}$} \\ \textbf{Accuracy}} & \makecell{\textbf{$\text{Wood Glue}^\text{K}$} \\ \textbf{Accuracy}} & \makecell{\textbf{$\text{Body Double}^\text{K}$} \\ \textbf{Accuracy}} \\ \midrule

\makecell{ResNet \\ DenseNet \\ Inception} & 0.9 & \makecell{0.908±0.009 \\ 0.931±0.013 \\ 0.959±0.002} & \makecell{0.819±0.002 \\ 0.838±0.022 \\ 0.879±0.006} & \makecell{0.927±0.009 \\ 0.949±0.013 \\ 0.966±0.002} & \makecell{0.703±0.006 \\ 0.717±0.050 \\ 0.786±0.012} & \makecell{6 \\ 5-7 \\ 4-5} & \makecell{1.881±0.074 \\ 2.111±0.145 \\ 2.462±0.036} & \makecell{0.595±0.024 \\ 0.511±0.046 \\ 0.379±0.039} & \makecell{0.591±0.010 \\ 0.600±0.060 \\ 0.701±0.024} & \makecell{0.823±0.012 \\ 0.837±0.043 \\ 0.870±0.004} & \makecell{0.643±0.016 \\ 0.655±0.056 \\ 0.733±0.014} & \makecell{0.779±0.027 \\ 0.775±0.068 \\ 0.865±0.013} & \makecell{0.788±0.008 \\ 0.829±0.015 \\ 0.846±0.015} \\
\bottomrule
\end{tabular}%
}

\vspace{0.5em} 

\caption{\textbf{Results for Scenario S3 using human-annotated AOI saliency}. Models trained using CYBORG loss in the limited-data context. Averaged over three runs.}

\end{table}

\begin{table}[H]
\centering

\label{tab:appendix-scenario3-human-boi-results}

\resizebox{\textwidth}{!}{%
\begin{tabular}{@{}l|c|c|c|c|c|c|c|c|c|c|c|c|c@{}}
\toprule
\textbf{Network} & \textbf{Alpha ($\alpha$)} & \textbf{AUC} & \textbf{Accuracy} & \makecell{\textbf{Bonafide} \\ \textbf{Accuracy}} & \makecell{\textbf{Spoof} \\ \textbf{Accuracy}} & \textbf{Placement} & \textbf{d'} & \makecell{\textbf{FNR @} \\ \textbf{FPR = 1\%}} & \makecell{\textbf{$\text{GLS20}^\text{U}$} \\ \textbf{Accuracy}} & \makecell{\textbf{$\text{VBN}^\text{U}$} \\ \textbf{Accuracy}} & \makecell{\textbf{$\text{RFastPro}^\text{U}$} \\ \textbf{Accuracy}} & \makecell{\textbf{$\text{Wood Glue}^\text{K}$} \\ \textbf{Accuracy}} & \makecell{\textbf{$\text{Body Double}^\text{K}$} \\ \textbf{Accuracy}} \\ \midrule

\makecell{ResNet \\ DenseNet \\ Inception} & 0.9 & \makecell{0.914±0.007 \\ 0.940±0.001 \\ 0.953±0.004} & \makecell{0.830±0.005 \\ 0.860±0.010 \\ 0.868±0.006} & \makecell{0.937±0.007 \\ 0.951±0.010 \\ 0.961±0.005} & \makecell{0.714±0.006 \\ 0.762±0.031 \\ 0.769±0.016} & \makecell{5 \\ 5 \\ 4-5} & \makecell{1.932±0.061 \\ 2.198±0.015 \\ 2.376±0.054} & \makecell{0.576±0.025 \\ 0.466±0.023 \\ 0.413±0.039} & \makecell{0.611±0.011 \\ 0.678±0.063 \\ 0.657±0.043} & \makecell{0.806±0.022 \\ 0.850±0.008 \\ 0.866±0.032} & \makecell{0.651±0.020 \\ 0.708±0.062 \\ 0.711±0.050} & \makecell{0.769±0.015 \\ 0.805±0.015 \\ 0.855±0.048} & \makecell{0.833±0.007 \\ 0.851±0.003 \\ 0.858±0.016} \\
\bottomrule
\end{tabular}%
}

\vspace{0.5em} 

\caption{\textbf{Results for Scenario S3 using human-annotated BOI saliency}. Models trained using CYBORG loss in the limited-data context. Averaged over three runs.}

\end{table}

\begin{table}[H]
\centering

\label{tab:appendix-scenario3-minutiae-foi-results}

\resizebox{\textwidth}{!}{%
\begin{tabular}{@{}l|c|c|c|c|c|c|c|c|c|c|c|c|c@{}}
\toprule
\textbf{Network} & \textbf{Alpha ($\alpha$)} & \textbf{AUC} & \textbf{Accuracy} & \makecell{\textbf{Bonafide} \\ \textbf{Accuracy}} & \makecell{\textbf{Spoof} \\ \textbf{Accuracy}} & \textbf{Placement} & \textbf{d'} & \makecell{\textbf{FNR @} \\ \textbf{FPR = 1\%}} & \makecell{\textbf{$\text{GLS20}^\text{U}$} \\ \textbf{Accuracy}} & \makecell{\textbf{$\text{VBN}^\text{U}$} \\ \textbf{Accuracy}} & \makecell{\textbf{$\text{RFastPro}^\text{U}$} \\ \textbf{Accuracy}} & \makecell{\textbf{$\text{Wood Glue}^\text{K}$} \\ \textbf{Accuracy}} & \makecell{\textbf{$\text{Body Double}^\text{K}$} \\ \textbf{Accuracy}} \\ \midrule

\makecell{ResNet \\ DenseNet \\ Inception} & 0.9 & \makecell{0.924±0.014 \\ 0.942±0.012 \\ 0.955±0.002} & \makecell{0.835±0.017 \\ 0.861±0.008 \\ 0.875±0.003} & \makecell{0.925±0.008 \\ 0.948±0.009 \\ 0.965±0.002} & \makecell{0.738±0.032 \\ 0.768±0.011 \\ 0.777±0.007} & \makecell{5-7 \\ 5 \\ 4-5} & \makecell{2.031±0.132 \\ 2.236±0.148 \\ 2.396±0.028} & \makecell{0.593±0.040 \\ 0.468±0.050 \\ 0.387±0.011} & \makecell{0.642±0.050 \\ 0.689±0.011 \\ 0.696±0.019} & \makecell{0.827±0.032 \\ 0.846±0.021 \\ 0.834±0.036} & \makecell{0.684±0.056 \\ 0.723±0.015 \\ 0.734±0.015} & \makecell{0.808±0.045 \\ 0.824±0.032 \\ 0.851±0.039} & \makecell{0.821±0.023 \\ 0.833±0.016 \\ 0.844±0.020} \\
\bottomrule
\end{tabular}%
}

\vspace{0.5em} 

\caption{\textbf{Results for Scenario S3 using minutiae-based FOI saliency}. Models trained using CYBORG loss in the limited-data context. Averaged over three runs.}

\end{table}

\begin{table}[H]
\centering

\label{tab:appendix-scenario3-minutiae-aoi-results}

\resizebox{\textwidth}{!}{%
\begin{tabular}{@{}l|c|c|c|c|c|c|c|c|c|c|c|c|c@{}}
\toprule
\textbf{Network} & \textbf{Alpha ($\alpha$)} & \textbf{AUC} & \textbf{Accuracy} & \makecell{\textbf{Bonafide} \\ \textbf{Accuracy}} & \makecell{\textbf{Spoof} \\ \textbf{Accuracy}} & \textbf{Placement} & \textbf{d'} & \makecell{\textbf{FNR @} \\ \textbf{FPR = 1\%}} & \makecell{\textbf{$\text{GLS20}^\text{U}$} \\ \textbf{Accuracy}} & \makecell{\textbf{$\text{VBN}^\text{U}$} \\ \textbf{Accuracy}} & \makecell{\textbf{$\text{RFastPro}^\text{U}$} \\ \textbf{Accuracy}} & \makecell{\textbf{$\text{Wood Glue}^\text{K}$} \\ \textbf{Accuracy}} & \makecell{\textbf{$\text{Body Double}^\text{K}$} \\ \textbf{Accuracy}} \\ \midrule

\makecell{ResNet \\ DenseNet \\ Inception} & 0.9 & \makecell{0.930±0.014 \\ 0.933±0.005 \\ 0.956±0.007} & \makecell{0.853±0.022 \\ 0.839±0.023 \\ 0.881±0.013} & \makecell{0.927±0.006 \\ 0.946±0.010 \\ 0.957±0.001} & \makecell{0.774±0.048 \\ 0.724±0.057 \\ 0.800±0.027} & \makecell{5-6 \\ 5-7 \\ 4-5} & \makecell{2.097±0.141 \\ 2.120±0.053 \\ 2.427±0.108} & \makecell{0.520±0.047 \\ 0.530±0.018 \\ 0.416±0.102} & \makecell{0.687±0.053 \\ 0.632±0.068 \\ 0.705±0.055} & \makecell{0.860±0.048 \\ 0.794±0.064 \\ 0.893±0.033} & \makecell{0.731±0.054 \\ 0.678±0.069 \\ 0.746±0.053} & \makecell{0.839±0.065 \\ 0.783±0.062 \\ 0.885±0.044} & \makecell{0.834±0.029 \\ 0.812±0.031 \\ 0.866±0.014} \\
\bottomrule
\end{tabular}%
}

\vspace{0.5em} 

\caption{\textbf{Results for Scenario S3 using minutiae-based AOI saliency}. Models trained using CYBORG loss in the limited-data context. Averaged over three runs.}

\end{table}

\begin{table}[H]
\centering

\label{tab:appendix-scenario3-minutiae-boi-results}

\resizebox{\textwidth}{!}{%
\begin{tabular}{@{}l|c|c|c|c|c|c|c|c|c|c|c|c|c@{}}
\toprule
\textbf{Network} & \textbf{Alpha ($\alpha$)} & \textbf{AUC} & \textbf{Accuracy} & \makecell{\textbf{Bonafide} \\ \textbf{Accuracy}} & \makecell{\textbf{Spoof} \\ \textbf{Accuracy}} & \textbf{Placement} & \textbf{d'} & \makecell{\textbf{FNR @} \\ \textbf{FPR = 1\%}} & \makecell{\textbf{$\text{GLS20}^\text{U}$} \\ \textbf{Accuracy}} & \makecell{\textbf{$\text{VBN}^\text{U}$} \\ \textbf{Accuracy}} & \makecell{\textbf{$\text{RFastPro}^\text{U}$} \\ \textbf{Accuracy}} & \makecell{\textbf{$\text{Wood Glue}^\text{K}$} \\ \textbf{Accuracy}} & \makecell{\textbf{$\text{Body Double}^\text{K}$} \\ \textbf{Accuracy}} \\ \midrule

\makecell{ResNet \\ DenseNet \\ Inception} & 0.9 & \makecell{0.912±0.006 \\ 0.929±0.004 \\ 0.951±0.003} & \makecell{0.828±0.012 \\ 0.841±0.006 \\ 0.867±0.001} & \makecell{0.925±0.009 \\ 0.945±0.014 \\ 0.961±0.008} & \makecell{0.724±0.015 \\ 0.728±0.020 \\ 0.765±0.009} & \makecell{5-6 \\ 5 \\ 5} & \makecell{1.912±0.054 \\ 2.079±0.042 \\ 2.338±0.046} & \makecell{0.576±0.049 \\ 0.509±0.024 \\ 0.430±0.037} & \makecell{0.606±0.032 \\ 0.628±0.053 \\ 0.665±0.051} & \makecell{0.839±0.027 \\ 0.790±0.061 \\ 0.857±0.037} & \makecell{0.654±0.026 \\ 0.688±0.051 \\ 0.715±0.037} & \makecell{0.820±0.028 \\ 0.765±0.076 \\ 0.849±0.042} & \makecell{0.816±0.023 \\ 0.845±0.009 \\ 0.835±0.016} \\
\bottomrule
\end{tabular}%
}

\vspace{0.5em} 

\caption{\textbf{Results for Scenario S3 using minutiae-based BOI saliency}. Models trained using CYBORG loss in the limited-data context. Averaged over three runs.}

\end{table}

\begin{table}[H]
\centering

\label{tab:appendix-scenario3-low-quality-results}

\resizebox{\textwidth}{!}{%
\begin{tabular}{@{}l|c|c|c|c|c|c|c|c|c|c|c|c|c@{}}
\toprule
\textbf{Network} & \textbf{Alpha ($\alpha$)} & \textbf{AUC} & \textbf{Accuracy} & \makecell{\textbf{Bonafide} \\ \textbf{Accuracy}} & \makecell{\textbf{Spoof} \\ \textbf{Accuracy}} & \textbf{Placement} & \textbf{d'} & \makecell{\textbf{FNR @} \\ \textbf{FPR = 1\%}} & \makecell{\textbf{$\text{GLS20}^\text{U}$} \\ \textbf{Accuracy}} & \makecell{\textbf{$\text{VBN}^\text{U}$} \\ \textbf{Accuracy}} & \makecell{\textbf{$\text{RFastPro}^\text{U}$} \\ \textbf{Accuracy}} & \makecell{\textbf{$\text{Wood Glue}^\text{K}$} \\ \textbf{Accuracy}} & \makecell{\textbf{$\text{Body Double}^\text{K}$} \\ \textbf{Accuracy}} \\ \midrule

\makecell{ResNet \\ DenseNet \\ Inception} & 0.9 & \makecell{0.931±0.002 \\ 0.931±0.006 \\ 0.955±0.004} & \makecell{0.836±0.010 \\ 0.846±0.013 \\ 0.873±0.005} & \makecell{0.941±0.004 \\ 0.940±0.009 \\ 0.957±0.010} & \makecell{0.722±0.019 \\ 0.745±0.036 \\ 0.782±0.009} & \makecell{5-6 \\ 5 \\ 4-5} & \makecell{2.093±0.017 \\ 2.102±0.067 \\ 2.404±0.068} & \makecell{0.582±0.016 \\ 0.529±0.024 \\ 0.441±0.078} & \makecell{0.603±0.024 \\ 0.650±0.044 \\ 0.698±0.007} & \makecell{0.862±0.018 \\ 0.834±0.040 \\ 0.866±0.022} & \makecell{0.642±0.020 \\ 0.696±0.044 \\ 0.740±0.018} & \makecell{0.807±0.030 \\ 0.808±0.054 \\ 0.820±0.015} & \makecell{0.822±0.018 \\ 0.828±0.010 \\ 0.862±0.005} \\
\bottomrule
\end{tabular}%
}

\vspace{0.5em} 

\caption{\textbf{Results for Scenario S3 using low-quality map saliency}. Models trained using CYBORG loss in the limited-data context. Averaged over three runs.}

\end{table}

\section{Scenario S4, Full Results}

\begin{table}[H]
\centering

\label{tab:appendix-scenario4-blurring-results}

\resizebox{\textwidth}{!}{%
\begin{tabular}{@{}l|c|c|c|c|c|c|c|c|c|c|c|c|c@{}}
\toprule
\textbf{Network} & \textbf{Saliency} & \textbf{AUC} & \textbf{Accuracy} & \makecell{\textbf{Bonafide} \\ \textbf{Accuracy}} & \makecell{\textbf{Spoof} \\ \textbf{Accuracy}} & \textbf{Placement} & \textbf{d'} & \makecell{\textbf{FNR @} \\ \textbf{FPR = 1\%}} & \makecell{\textbf{$\text{GLS20}^\text{U}$} \\ \textbf{Accuracy}} & \makecell{\textbf{$\text{VBN}^\text{U}$} \\ \textbf{Accuracy}} & \makecell{\textbf{$\text{RFastPro}^\text{U}$} \\ \textbf{Accuracy}} & \makecell{\textbf{$\text{Wood Glue}^\text{K}$} \\ \textbf{Accuracy}} & \makecell{\textbf{$\text{Body Double}^\text{K}$} \\ \textbf{Accuracy}} \\ \midrule

\makecell{ResNet \\ DenseNet \\ Inception} & None & \makecell{0.869±0.015 \\ 0.846±0.027 \\ 0.878±0.013} & \makecell{0.786±0.011 \\ 0.762±0.023 \\ 0.794±0.018} & \makecell{0.876±0.017 \\ 0.837±0.004 \\ 0.869±0.019} & \makecell{0.689±0.020 \\ 0.681±0.049 \\ 0.714±0.057} & \makecell{7-8 \\ 7-12 \\ 6-9} & \makecell{1.593±0.100 \\ 1.452±0.158 \\ 1.652±0.094} & \makecell{0.769±0.037 \\ 0.785±0.008 \\ 0.741±0.030}  & \makecell{0.668±0.006 \\ 0.660±0.028 \\ 0.713±0.039} & \makecell{0.707±0.043 \\ 0.712±0.109 \\ 0.691±0.094} & \makecell{0.709±0.007 \\ 0.705±0.037 \\ 0.758±0.042} & \makecell{0.623±0.048 \\ 0.598±0.095 \\ 0.588±0.123} & \makecell{0.729±0.055 \\ 0.721±0.019 \\ 0.773±0.031} \\
\bottomrule
\end{tabular}%
}

\vspace{0.5em} 

\caption{\textbf{Results for Scenario S4, guided by blurring non-salient regions}. Models trained using cross-entropy loss in the limited-data context. Averaged over three runs.}

\end{table}

\section{Scenario S5, Full Results}

\begin{table}[H]
\centering

\label{tab:appendix-scenario5-autoencoder-foi-results}

\resizebox{\textwidth}{!}{%
\begin{tabular}{@{}l|c|c|c|c|c|c|c|c|c|c|c|c|c@{}}
\toprule
\textbf{Network} & \textbf{Alpha ($\alpha$)} & \textbf{AUC} & \textbf{Accuracy} & \makecell{\textbf{Bonafide} \\ \textbf{Accuracy}} & \makecell{\textbf{Spoof} \\ \textbf{Accuracy}} & \textbf{Placement} & \textbf{d'} & \makecell{\textbf{FNR @} \\ \textbf{FPR = 1\%}} & \makecell{\textbf{$\text{GLS20}^\text{U}$} \\ \textbf{Accuracy}} & \makecell{\textbf{$\text{VBN}^\text{U}$} \\ \textbf{Accuracy}} & \makecell{\textbf{$\text{RFastPro}^\text{U}$} \\ \textbf{Accuracy}} & \makecell{\textbf{$\text{Wood Glue}^\text{K}$} \\ \textbf{Accuracy}} & \makecell{\textbf{$\text{Body Double}^\text{K}$} \\ \textbf{Accuracy}} \\ \midrule

\makecell{ResNet \\ DenseNet \\ Inception} & 0.9 & \makecell{0.988±0.001 \\ 0.982±0.001 \\ 0.991±0.001} & \makecell{0.900±0.007 \\ 0.894±0.007 \\ 0.911±0.016} & \makecell{0.996±0.001 \\ 0.991±0.003 \\ 0.995±0.003} & \makecell{0.795±0.016 \\ 0.790±0.017 \\ 0.822±0.037} & \makecell{4 \\ 4 \\ 2-4} & \makecell{3.199±0.065 \\ 2.978±0.020 \\ 3.328±0.075} & \makecell{0.141±0.008 \\ 0.201±0.017 \\ 0.132±0.008}  & \makecell{0.704±0.026 \\ 0.685±0.026 \\ 0.738±0.055} & \makecell{0.897±0.002 \\ 0.909±0.007 \\ 0.905±0.014} & \makecell{0.727±0.024 \\ 0.716±0.022 \\ 0.751±0.061} & \makecell{0.886±0.016 \\ 0.919±0.017 \\ 0.912±0.009} & \makecell{0.864±0.005 \\ 0.838±0.016 \\ 0.898±0.028} \\
\bottomrule
\end{tabular}%
}

\vspace{0.5em} 

\caption{\textbf{Results for Scenario S5 using autoencoder-generated FOI saliency}. Models trained using CYBORG in the large-data-availability context. Averaged over three runs.}

\end{table}

\begin{table}[H]
\centering

\label{tab:appendix-scenario5-autoencoder-aoi-results}

\resizebox{\textwidth}{!}{%
\begin{tabular}{@{}l|c|c|c|c|c|c|c|c|c|c|c|c|c@{}}
\toprule
\textbf{Network} & \textbf{Alpha ($\alpha$)} & \textbf{AUC} & \textbf{Accuracy} & \makecell{\textbf{Bonafide} \\ \textbf{Accuracy}} & \makecell{\textbf{Spoof} \\ \textbf{Accuracy}} & \textbf{Placement} & \textbf{d'} & \makecell{\textbf{FNR @} \\ \textbf{FPR = 1\%}} & \makecell{\textbf{$\text{GLS20}^\text{U}$} \\ \textbf{Accuracy}} & \makecell{\textbf{$\text{VBN}^\text{U}$} \\ \textbf{Accuracy}} & \makecell{\textbf{$\text{RFastPro}^\text{U}$} \\ \textbf{Accuracy}} & \makecell{\textbf{$\text{Wood Glue}^\text{K}$} \\ \textbf{Accuracy}} & \makecell{\textbf{$\text{Body Double}^\text{K}$} \\ \textbf{Accuracy}} \\ \midrule

\makecell{ResNet \\ DenseNet \\ Inception} & 0.9 & \makecell{0.987±0.000 \\ 0.985±0.003 \\ 0.990±0.001} & \makecell{0.904±0.008 \\ 0.899±0.015 \\ 0.916±0.011} & \makecell{0.995±0.001 \\ 0.993±0.002 \\ 0.994±0.001} & \makecell{0.806±0.017 \\ 0.798±0.033 \\ 0.831±0.024} & \makecell{3-4 \\ 4 \\ 2-4} & \makecell{3.141±0.018 \\ 3.092±0.122 \\ 3.317±0.084} & \makecell{0.154±0.018 \\ 0.179±0.019 \\ 0.131±0.008}  & \makecell{0.719±0.011 \\ 0.694±0.037 \\ 0.742±0.033} & \makecell{0.904±0.018 \\ 0.909±0.021 \\ 0.931±0.012} & \makecell{0.747±0.015 \\ 0.727±0.043 \\ 0.765±0.029} & \makecell{0.904±0.020 \\ 0.915±0.022 \\ 0.938±0.015} & \makecell{0.854±0.027 \\ 0.859±0.034 \\ 0.881±0.030} \\
\bottomrule
\end{tabular}%
}

\vspace{0.5em} 

\caption{\textbf{Results for Scenario S5 using autoencoder-generated AOI saliency}. Models trained using CYBORG in the large-data-availability context. Averaged over three runs.}

\end{table}

\begin{table}[H]
\centering

\label{tab:appendix-scenario5-autoencoder-boi-results}

\resizebox{\textwidth}{!}{%
\begin{tabular}{@{}l|c|c|c|c|c|c|c|c|c|c|c|c|c@{}}
\toprule
\textbf{Network} & \textbf{Alpha ($\alpha$)} & \textbf{AUC} & \textbf{Accuracy} & \makecell{\textbf{Bonafide} \\ \textbf{Accuracy}} & \makecell{\textbf{Spoof} \\ \textbf{Accuracy}} & \textbf{Placement} & \textbf{d'} & \makecell{\textbf{FNR @} \\ \textbf{FPR = 1\%}} & \makecell{\textbf{$\text{GLS20}^\text{U}$} \\ \textbf{Accuracy}} & \makecell{\textbf{$\text{VBN}^\text{U}$} \\ \textbf{Accuracy}} & \makecell{\textbf{$\text{RFastPro}^\text{U}$} \\ \textbf{Accuracy}} & \makecell{\textbf{$\text{Wood Glue}^\text{K}$} \\ \textbf{Accuracy}} & \makecell{\textbf{$\text{Body Double}^\text{K}$} \\ \textbf{Accuracy}} \\ \midrule

\makecell{ResNet \\ DenseNet \\ Inception} & 0.9 & \makecell{0.989±0.001 \\ 0.990±0.002 \\ 0.989±0.001} & \makecell{0.911±0.003 \\ 0.924±0.001 \\ 0.904±0.008} & \makecell{0.996±0.000 \\ 0.991±0.001 \\ 0.997±0.000} & \makecell{0.819±0.005 \\ 0.850±0.003 \\ 0.804±0.017} & \makecell{4 \\ 3 \\ 4} & \makecell{3.222±0.050 \\ 3.312±0.115 \\ 3.243±0.055} & \makecell{0.140±0.008 \\ 0.139±0.007 \\ 0.134±0.011}  & \makecell{0.745±0.010 \\ 0.776±0.014 \\ 0.715±0.022} & \makecell{0.889±0.005 \\ 0.927±0.021 \\ 0.900±0.018} & \makecell{0.777±0.017 \\ 0.804±0.016 \\ 0.747±0.014} & \makecell{0.886±0.006 \\ 0.931±0.022 \\ 0.905±0.013} & \makecell{0.870±0.003 \\ 0.892±0.008 \\ 0.851±0.030} \\
\bottomrule
\end{tabular}%
}

\vspace{0.5em} 

\caption{\textbf{Results for Scenario S5 using autoencoder-generated BOI saliency}. Models trained using CYBORG in the large-data-availability context. Averaged over three runs.}

\end{table}

\begin{table}[H]
\centering

\label{tab:appendix-scenario5-minutiae-foi-results}

\resizebox{\textwidth}{!}{%
\begin{tabular}{@{}l|c|c|c|c|c|c|c|c|c|c|c|c|c@{}}
\toprule
\textbf{Network} & \textbf{Alpha ($\alpha$)} & \textbf{AUC} & \textbf{Accuracy} & \makecell{\textbf{Bonafide} \\ \textbf{Accuracy}} & \makecell{\textbf{Spoof} \\ \textbf{Accuracy}} & \textbf{Placement} & \textbf{d'} & \makecell{\textbf{FNR @} \\ \textbf{FPR = 1\%}} & \makecell{\textbf{$\text{GLS20}^\text{U}$} \\ \textbf{Accuracy}} & \makecell{\textbf{$\text{VBN}^\text{U}$} \\ \textbf{Accuracy}} & \makecell{\textbf{$\text{RFastPro}^\text{U}$} \\ \textbf{Accuracy}} & \makecell{\textbf{$\text{Wood Glue}^\text{K}$} \\ \textbf{Accuracy}} & \makecell{\textbf{$\text{Body Double}^\text{K}$} \\ \textbf{Accuracy}} \\ \midrule

\makecell{ResNet \\ DenseNet \\ Inception} & 0.9 & \makecell{0.985±0.002 \\ 0.988±0.002 \\ 0.989±0.002} & \makecell{0.899±0.003 \\ 0.922±0.008 \\ 0.914±0.001} & \makecell{0.994±0.002 \\ 0.989±0.003 \\ 0.995±0.000} & \makecell{0.795±0.004 \\ 0.849±0.019 \\ 0.826±0.001} & \makecell{4 \\ 2-3 \\ 3-4} & \makecell{3.079±0.089 \\ 3.179±0.076 \\ 3.235±0.097} & \makecell{0.161±0.013 \\ 0.153±0.013 \\ 0.132±0.005}  & \makecell{0.708±0.011 \\ 0.774±0.027 \\ 0.731±0.005} & \makecell{0.875±0.031 \\ 0.893±0.018 \\ 0.920±0.008} & \makecell{0.738±0.014 \\ 0.807±0.018 \\ 0.765±0.003} & \makecell{0.893±0.016 \\ 0.937±0.012 \\ 0.919±0.016} & \makecell{0.855±0.006 \\ 0.903±0.022 \\ 0.892±0.005} \\
\bottomrule
\end{tabular}%
}

\vspace{0.5em} 

\caption{\textbf{Results for Scenario S5 using minutiae-based FOI saliency}. Models trained using CYBORG in the large-data-availability context. Averaged over three runs.}

\end{table}

\begin{table}[H]
\centering

\label{tab:appendix-scenario5-minutiae-aoi-results}

\resizebox{\textwidth}{!}{%
\begin{tabular}{@{}l|c|c|c|c|c|c|c|c|c|c|c|c|c@{}}
\toprule
\textbf{Network} & \textbf{Alpha ($\alpha$)} & \textbf{AUC} & \textbf{Accuracy} & \makecell{\textbf{Bonafide} \\ \textbf{Accuracy}} & \makecell{\textbf{Spoof} \\ \textbf{Accuracy}} & \textbf{Placement} & \textbf{d'} & \makecell{\textbf{FNR @} \\ \textbf{FPR = 1\%}} & \makecell{\textbf{$\text{GLS20}^\text{U}$} \\ \textbf{Accuracy}} & \makecell{\textbf{$\text{VBN}^\text{U}$} \\ \textbf{Accuracy}} & \makecell{\textbf{$\text{RFastPro}^\text{U}$} \\ \textbf{Accuracy}} & \makecell{\textbf{$\text{Wood Glue}^\text{K}$} \\ \textbf{Accuracy}} & \makecell{\textbf{$\text{Body Double}^\text{K}$} \\ \textbf{Accuracy}} \\ \midrule

\makecell{ResNet \\ DenseNet \\ Inception} & 0.9 & \makecell{0.989±0.001 \\ 0.988±0.001 \\ 0.990±0.002} & \makecell{0.915±0.002 \\ 0.938±0.004 \\ 0.916±0.010} & \makecell{0.996±0.002 \\ 0.964±0.001 \\ 0.993±0.001} & \makecell{0.829±0.004 \\ 0.910±0.009 \\ 0.832±0.021} & \makecell{3-4 \\ 1-2 \\ 3-4} & \makecell{3.222±0.049 \\ 3.175±0.058 \\ 3.308±0.093} & \makecell{0.134±0.009 \\ 0.167±0.010 \\ 0.144±0.012} & \makecell{0.750±0.012 \\ 0.857±0.009 \\ 0.745±0.027} & \makecell{0.900±0.001 \\ 0.927±0.023 \\ 0.915±0.011} & \makecell{0.782±0.006 \\ 0.887±0.010 \\ 0.786±0.041} & \makecell{0.909±0.007 \\ 0.974±0.006 \\ 0.928±0.024} & \makecell{0.885±0.006 \\ 0.949±0.009 \\ 0.875±0.010} \\
\bottomrule
\end{tabular}%
}

\vspace{0.5em} 

\caption{\textbf{Results for Scenario S5 using minutiae-based AOI saliency}. Models trained using CYBORG in the large-data-availability context. Averaged over three runs.}

\end{table}

\begin{table}[H]
\centering

\label{tab:appendix-scenario5-minutiae-boi-results}

\resizebox{\textwidth}{!}{%
\begin{tabular}{@{}l|c|c|c|c|c|c|c|c|c|c|c|c|c@{}}
\toprule
\textbf{Network} & \textbf{Alpha ($\alpha$)} & \textbf{AUC} & \textbf{Accuracy} & \makecell{\textbf{Bonafide} \\ \textbf{Accuracy}} & \makecell{\textbf{Spoof} \\ \textbf{Accuracy}} & \textbf{Placement} & \textbf{d'} & \makecell{\textbf{FNR @} \\ \textbf{FPR = 1\%}} & \makecell{\textbf{$\text{GLS20}^\text{U}$} \\ \textbf{Accuracy}} & \makecell{\textbf{$\text{VBN}^\text{U}$} \\ \textbf{Accuracy}} & \makecell{\textbf{$\text{RFastPro}^\text{U}$} \\ \textbf{Accuracy}} & \makecell{\textbf{$\text{Wood Glue}^\text{K}$} \\ \textbf{Accuracy}} & \makecell{\textbf{$\text{Body Double}^\text{K}$} \\ \textbf{Accuracy}} \\ \midrule

\makecell{ResNet \\ DenseNet \\ Inception} & 0.9 & \makecell{0.985±0.003 \\ 0.930±0.026 \\ 0.931±0.031} & \makecell{0.903±0.007 \\ 0.847±0.020 \\ 0.862±0.029} & \makecell{0.996±0.001 \\ 0.919±0.033 \\ 0.962±0.017} & \makecell{0.803±0.014 \\ 0.769±0.007 \\ 0.754±0.042} & \makecell{4 \\ 5-6 \\ 4-5} & \makecell{3.090±0.128 \\ 2.123±0.274 \\ 2.177±0.397} & \makecell{0.150±0.016 \\ 0.415±0.022 \\ 0.347±0.098} & \makecell{0.711±0.030 \\ 0.632±0.003 \\ 0.649±0.053} & \makecell{0.891±0.001 \\ 0.919±0.015 \\ 0.856±0.011} & \makecell{0.739±0.025 \\ 0.684±0.004 \\ 0.709±0.057} & \makecell{0.901±0.006 \\ 0.909±0.017 \\ 0.866±0.014} & \makecell{0.872±0.008 \\ 0.846±0.021 \\ 0.789±0.057} \\
\bottomrule
\end{tabular}%
}

\vspace{0.5em} 

\caption{\textbf{Results for Scenario S5 using minutiae-based BOI saliency}. Models trained using CYBORG in the large-data-availability context. Averaged over three runs.}

\end{table}

\begin{table}[H]
\centering

\label{tab:appendix-scenario5-low-quality-results}

\resizebox{\textwidth}{!}{%
\begin{tabular}{@{}l|c|c|c|c|c|c|c|c|c|c|c|c|c@{}}
\toprule
\textbf{Network} & \textbf{Alpha ($\alpha$)} & \textbf{AUC} & \textbf{Accuracy} & \makecell{\textbf{Bonafide} \\ \textbf{Accuracy}} & \makecell{\textbf{Spoof} \\ \textbf{Accuracy}} & \textbf{Placement} & \textbf{d'} & \makecell{\textbf{FNR @} \\ \textbf{FPR = 1\%}} & \makecell{\textbf{$\text{GLS20}^\text{U}$} \\ \textbf{Accuracy}} & \makecell{\textbf{$\text{VBN}^\text{U}$} \\ \textbf{Accuracy}} & \makecell{\textbf{$\text{RFastPro}^\text{U}$} \\ \textbf{Accuracy}} & \makecell{\textbf{$\text{Wood Glue}^\text{K}$} \\ \textbf{Accuracy}} & \makecell{\textbf{$\text{Body Double}^\text{K}$} \\ \textbf{Accuracy}} \\ \midrule

\makecell{ResNet \\ DenseNet \\ Inception} & 0.9 & \makecell{0.983±0.003 \\ 0.961±0.007 \\ 0.991±0.002} & \makecell{0.895±0.005 \\ 0.871±0.015 \\ 0.906±0.011} & \makecell{0.994±0.001 \\ 0.966±0.003 \\ 0.996±0.001} & \makecell{0.788±0.011 \\ 0.768±0.027 \\ 0.808±0.023} & \makecell{4 \\ 4-5 \\ 3-4} & \makecell{2.995±0.101 \\ 2.507±0.127 \\ 3.328±0.090} & \makecell{0.170±0.008 \\ 0.555±0.048 \\ 0.138±0.012}  & \makecell{0.709±0.019 \\ 0.656±0.048 \\ 0.710±0.031} & \makecell{0.885±0.003 \\ 0.903±0.012 \\ 0.922±0.006} & \makecell{0.727±0.016 \\ 0.702±0.037 \\ 0.734±0.038} & \makecell{0.887±0.009 \\ 0.924±0.006 \\ 0.933±0.003} & \makecell{0.826±0.008 \\ 0.780±0.021 \\ 0.859±0.027} \\
\bottomrule
\end{tabular}%
}

\vspace{0.5em} 

\caption{\textbf{Results for Scenario S5 using low-quality map saliency}. Models trained using CYBORG in the large-data-availability context. Averaged over three runs.}

\end{table}

\end{document}